\documentclass[11pt,draftcls,onecolumn]{IEEEtran}
\usepackage[cmex10]{amsmath}
\usepackage{adjustbox}
\usepackage{tabularx}
\usepackage{subfig}
\usepackage{array}
\newcolumntype{L}{>{\centering\arraybackslash}m{3cm}}
\usepackage{multicol}
\usepackage{tabularx}
\usepackage{enumerate}
\usepackage{graphicx}
\usepackage{float}
\usepackage{booktabs}
\usepackage{color}
\usepackage{amssymb}
\usepackage{epstopdf}
\UseRawInputEncoding
\DeclareGraphicsExtensions{.eps,.pdf,.png}

\usepackage[switch]{lineno}
\begin{document}
\setlength{\parskip}{0pt}
\title{Does Explainable Machine Learning Uncover the Black Box in Vision Applications?}

\author{Manish Narwaria$^*$
\thanks{$^*$Department of Electrical Engineering, Indian Institute of Technology Jodhpur, NH 62, Surpura Bypass Rd, Karwar, Rajasthan 342037, India. e-mail: narwaria@iitj.ac.in} 
\thanks{}}

\maketitle

\begin{abstract}

Machine learning (ML) in general and deep learning (DL) in particular has become an extremely popular tool in several vision applications (like object detection, super resolution, segmentation, object tracking etc.). Almost in parallel, the issue of explainability in ML (i.e. the ability to explain/elaborate the way a trained ML model arrived at its decision) in vision has also received fairly significant attention from various quarters. However, we argue that the current philosophy behind explainable ML suffers from certain limitations, and the resulting explanations may not meaningfully uncover black box ML models. To elaborate our assertion, we first raise a few fundamental questions which have not been adequately discussed in the corresponding literature. We also provide perspectives on how explainablity in ML can benefit by relying on more rigorous principles in the related areas.
\end{abstract}

\section{The Need to Revisit Current Explainable ML Philosophy}

The powerful modeling capabilities of ML/DL has fueled a transition from white box to black box modeling, in research, industry and education \cite{9418577}. Consequently, despite the success of ML/DL in several applications, the issue of explainability (i.e. ability to understand how an ML model arrives at a decision/prediction) remains one of the bottlenecks in adopting ML/DL models to a wider canvas of applications. Accordingly, it is fairly well accepted that ML/DL models should be amenable to a level of scrutiny that goes beyond simple audit. Hence, explainable ML remains an active research area \cite{xie2020explainable}, \cite{buhrmester2019analysis} in both industry and academia. However, there appears a lack of discussion on whether the current philosophy behind explainable ML can help to achieve the eventual goal of uncovering black box models. This is especially required in vision applications as meaningful visualization/explanation can greatly facilitate more transparent deployment of ML/DL models in practice. In that context, we raise three important questions which in our opinion are fundamental to a rigorous discussion on explainable ML, and have not been adequately raised or addressed in the current literature. We also attempt to provide perspectives and some insights into these questions. Hence, the goal is not to criticize existing efforts on explainable ML. Rather it is to build upon them, and in the process raise awareness about possible issues whose mitigation may help to develop more understandable ML/DL models.
\subsection{Explainable ML: A Prehoc Necessity or A Posthoc Evil?} \label{need}
A commonly accepted argument for the need of explainability in ML is based on how serious the consequences might be if the trained model makes an error. For instance, consider medical image analysis where an ML/DL algorithm erroneously classifies an image as ``normal'' when in fact there was say a tumor (or some other serious medical abnormality) present. Obviously, such an error can have serious medical implications. Likewise, errors made by an ML algorithm in the context of autonomous vehicles might also undeniably lead to disastrous consequences. On the other hand, errors made by ML/DL are considered less serious, from a practical view point, in several other vision applications such as object recognition in photos/videos on the web, visual content retrieval, gaming, vision based recommender systems, audiovisual communication etc. Such a dichotomous but seemingly logical philosophy has unfortunately lead to a proliferation of black box models in the first place. As a result, explainability has been possibly relegated to merely a \emph{posthoc} analysis of the trained ML model rather than being be viewed as a fundamental and inherent concept in ML/DL algorithm design. It is therefore reasonable to ask if we should revisit the current philosophy by making explainability a pre-design necessity and not merely a post design baggage.  

\subsection{Do visual explanations generate new knowledge?} \label{knowledge}
\begin{figure}[!t]
\center
\includegraphics[scale=0.58]{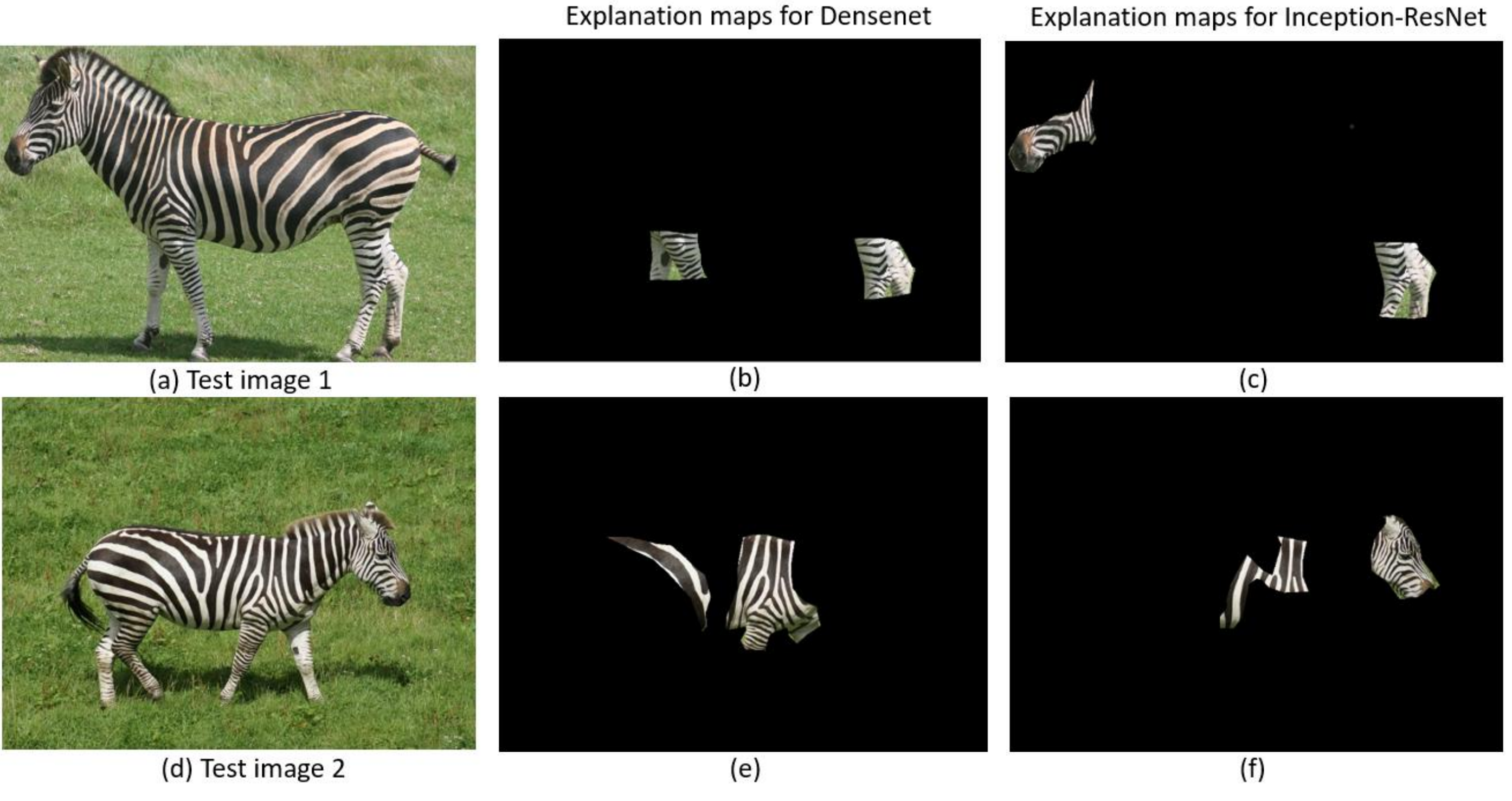}
\caption{Visual explanations generated for two test images. The explanations were generated by LIME method \cite{10.1145/2939672.2939778} for pre-trained Densenet and Inception-Resnet models. Only super pixels that correspond to top two features as determined by LIME are shown for easier visualization.}
\label{expl}
\end{figure}

DL in particular has gained popularity since it achieves better prediction performance than traditional non ML and ML approaches, especially in many vision applications. This can be largely attributed to the fact that feature extraction in DL is typically implicit (i.e. does not require handcrafted features) and data dependent. In contrast, traditional approaches rely heavily on feature engineering (explicit modeling) and possible use of apriori domain knowledge. It is therefore natural to ask: what additional information was extracted and exploited by the DL model that was missed in the more traditional approach for solving the same problem? Stated differently, can the explanations obtained from trained ML/DL models bridge the seeming gap in terms of additional knowledge that DL supposedly exploits which lead to better performances in vision applications? In the context of this question, we note that several existing efforts attempt to ``explain'' how DL works in applications like image classification/recognition etc. by analyzing important regions (in the form of superpixels \cite{10.1145/2939672.2939778}, salient regions \cite{DBLP:conf/bmvc/PetsiukDS18} or similar ideas like layer-wise relevance propagation LRP \cite{bach-plos15} etc.). Hence, the main idea is to find regions in a test image (video) on which the trained DL network focuses on (or has relevance) for the purpose of classification/recognition. This information presumably explains how the ML/DL algorithm arrives at a decision. To analyze this in more details, it is convenient to consider the example in Figure \ref{expl} where we used two test images in which the main object of interest is a zebra. We then employed two pre trained models namely Densenet \cite{8099726} and Inception-Resnet \cite{10.5555/3298023.3298188} to recognize the object in the test images. We found that both the models (pre-trained on nearly a million images from Imagnet \cite{5206848}) correctly recognized the object namely `zebra' (with a probability above 0.9) in both the test images. However, from an explainable ML view point, it will be obviously more interesting to understand visual signal information that the two DL models used to make the correct decision.  

To that end, we employed LIME \cite{10.1145/2939672.2939778} to get further insights into the two black boxes. LIME identifies regions in the image (in the form of superpixels) which were more important for the ML/DL algorithm to arrive at the decision. In Figure \ref{expl}, the second and third image in each row shows the two most important regions (visualized as explanation maps) that the respective DL models focused on to correctly classify the image as `zebra'. While the 4 explanation maps in Figure \ref{expl} (b), (c), (e) and (f) correspond to some parts of the object (zebra), two key questions remain unanswered. First, do these specific image parts stand out (i.e. are different) in comparison to other image regions and if so how? Second, are these parts amenable to interpretation in terms of simpler signals or patterns similar to basis functions (or basic building blocks) for the object under consideration (zebra)? 

\subsection{Can visual explanations constitute a many-to-one mapping?} \label{many-to-one}
Another important issue in explainable ML is that of a many-to-one mapping. It means that a unique object (class label) can be potentially mapped to more than one explanations. For example, `zebra' object in Figure \ref{expl} has 4 different explanations despite the fact that both the DL models recognized the object class correctly. In an analogous manner, we can conclude that a given DL model might not always rely on certain unique signal information (features) to correctly identify the same object. For instance, we observe from Figure \ref{expl} (b) that Densenet uses ``legs'' part of the zebra for first test image while it relies on middle body part (Figure \ref{expl} (e)) in case of the second test image. Likewise, while Inception-Resnet employs ``legs and face'' (Figure \ref{expl} (c)) region of the zebra for first test image but focuses on `` face and middle part of body'' (Figure \ref{expl} (f)) in the second test image. That is, the two DL models under consideration correctly identify the same object but rely on very different visual signal information in the two test images in Figure \ref{expl}. Moreover, the dimensionality of the mapped space (i.e. the number of explanations for the same object of interest) can increase quickly. Suppose we have $m$ images of an object of interest whose class (label) is being predicted by say $n$ ML models, and we employ $p$ explanation methods. Obviously, unless we have unique explanations, we are potentially looking at $m\times n\times p$ explanations for the same object. This leads to the following question: how logical is it that ML models correctly recognize the same object present in more than one test image but characterize it differently (i.e. use very different explanations for each test image to arrive at the correct prediction)?   

\section{Explainable ML: A Possible Way Forward}
We now attempt to provide some perspectives in the light of the questions raised. Before doing that, two points are however worth emphasizing. First, we note that the questions raised in the previous section are not influenced by our choice of specific test images in Figure \ref{expl} or the 2 DL models (Densenet and Inception-Resnet) or even the explanation method (LIME). Rather, the genesis of these questions lies primarily in the current philosophy behind explainable ML. Second, the said questions are not necessarily  independent of each other. Thus, it is possible that focusing on inherent ML model explainability right from the start may reveal useful insights about knowledge generation and uniqueness of explanations.

\subsection{Explainability as a first principle}
With regard to the question raised in section \ref{need}, it should be reasonable to conclude that explainability ought be an integral part of ML/DL algorithm design process itself and not an after thought. Thus, irrespective of the use-case scenario, an ML algorithm should be explainable/interpretable. Of course, the type of interpretability (i.e. the dimension) required can vary depending on the application (similar to how one views various tools in say signal processing). This could for instance involve learning of data-dependent but more intuitive and interpretable basis functions or representations. Likewise, ideas and design philosophy from signal processing, information theory and related disciplines can be borrowed. For example, fundamental ideas of frequency have been exploited to improve the interpretability of DL \cite{FTattributionpriors2020}. Other examples include the use of classical but interpretable concepts (such as filtering) in Digital Signal Processing (DSP) \cite{engel2020ddsp}, in vision \cite{CHELLAPPA20163} or from physics \cite{thuerey2021pbdl}, \cite{Kellman:EECS-2020-167} etc. Similarly, exploitation of apriori knowledge such as logical rules and knowledge graphs \cite{DBLP:journals/corr/abs-2105-10172} could lead to possibilities of enhancing our understanding of black box ML systems. Thus, it could be argued that irrespective of the approach eventually used, the design philosophy should focus on ML models that are inherently explainable. A \emph{posthoc} analysis could of course always be employed as an add-on for further refinement or seeking application specific insights.
\subsection{Possibility of Knowledge Generation and Transfer}
Supervised ML/DL approach depends heavily on data to derive mapping from a visual signal to the task of interest (eg. recognition of objects from images). Because the training process potentially uses a large amount of labeled data (millions of images for instance in Imagnet \cite{5206848}), it may not be unreasonable to expect that explanations of black box ML/DL models should reveal some new insights. However, it is likely that the current methods for explainablity, that tend to rely exclusively on posthoc analysis, are inadequate for this purpose. For instance, it is unclear if further analysis of the explanations (such as those in Figure \ref{expl}) say in terms of explicit shape, texture, orientation or color will reveal any new insights about the object under consideration. Thus, the answers to the questions discussed in section \ref{knowledge} are probably in the negative. But perhaps more pertinently, these questions are meant to emphasize the importance of explanations that are meaningfully quantifiable and uniquely interpretable at least in the context of specific application (or a set of related applications). Such functionality may eventually open the possibility of generating potentially new and explicit knowledge which is open to scrutiny and amenable to knowledge transfer. A relevant example in this regard would be that of vision science which includes calibrated psycho visual experiments and computational modeling \cite{10.5555/1095712}. The findings of such experiments are not only fundamental but have had tremendous impact on conceptualization and meaningful advancement of applications. This includes next generation video technologies like HFR \cite{7308124}, HDR \cite{CHALMERS201749}, visual saliency \cite{CARRASCO20111484}, modern video compression \cite{9146767}, to list a few. On similar lines,  a bottom-up and knowledge centric explainable ML design philosophy may be practically more useful and scalable.

\subsection{Uniqueness of explanations}
The keen reader will probably agree that the aspect of many-to-one mapping highlighted in section \ref{many-to-one} should in general be problematic in the context of explainable ML. Specifically, it implies that:
\begin{enumerate}
\item An ML/DL algorithm may not be using unique visual information to correctly identify the same object present in more than one test image.
\item The method used to explain/interpret black box models may not be capturing the correct features (information) that is being used by the ML model to make predictions.
\item Two or more ML models with same prediction accuracies might be using very different explanations. 
\end{enumerate}
The first point indicates lack of consistency and reliability of the ML/DL model under consideration. The second refers to possible deficiencies in the method itself that was used to explain the ML model. The third point is interesting since it refers to non equivalence of two or more ML models having the same prediction performance. That is, it would be more logical that a \emph{better} ML model should tend to use similar description (explanation) of the same object in more than one test image.  

Thus, consistency of explanations in addition to prediction accuracy should be a more reasonable performance benchmark metric. In an ideal case, one could hope that explanations for same object are same/similar across more than one test images. This may provide an opportunity to simply characterize those explanations and generate \emph{signatures} for different objects. In such case, deployment of ML models in practice would be greatly simplified since prediction could be done using logical rules and/or simpler models which in turn were derived from more complex DL/ML models trained on large sets of labeled data. Hence, uniqueness of explanations would increase ML model reliability and also open possibilities of new knowledge generation.

\section{Concluding Remarks}
There is no denying that ML models ought to be more transparent from the view point of signal information exploited to make predictions. We have, however, reasoned that the current philosophy behind explainable ML may not meaningfully uncover the black box. To that end, we highlighted three specific but not entirely independent aspects that probably deserve more attention. These include:  
\begin{enumerate}[\itshape i.]
\item A generally accepted but probably unnecessary dichotomy wherein the need for explainability depends on use-case scenario. As a result, explainability might not be treated as an inherent concept but merely an add-on in the context of ML algorithm design. 
\item An apparent lack of clarity on what the explanations imply and how they could help to improve our existing knowledge about specific tasks (say object recognition). The expectation about new knowledge generation should not be unreasonable especially when a tremendous effort goes into large scale data collection and labeling, and eventually training the model on modern computing devices.
\item The problem of many-to-one mapping where one class label (object) might be mapped to more than one explanations. Analogously, the same ML algorithm might use different features to identify the same object. 
\end{enumerate}
We also provided perspectives on the identified issues and possible ways of mitigating them. In particular, we emphasized that explainability should be accorded high priority in ML algorithm design process and not left merely as a \emph{posthoc} exercise. This could, for instance, be facilitated by relying on philosophy of fundamental principles in related domains. As already pointed out, such approach has already been employed \cite{FTattributionpriors2020}-\cite{DBLP:journals/corr/abs-2105-10172}, and could prove to be the gateway to meaningfully more transparent ML models. 
    
\bibliographystyle{IEEEtran}
\bibliography{publication}
%
%
%


\end{document}